\documentclass[11pt]{article}
\usepackage[utf8]{inputenc}
\usepackage[margin=0.97in]{geometry}
\usepackage{cite}
\usepackage{amsmath}
\usepackage{hyperref}
\usepackage{graphicx}

\title{The AI Scientific Community: Agentic Virtual Lab Swarms}
\author{Ulisses Braga-Neto\\
  Department of Electrical and Computer Engineering\\
Texas A\&M University}
\date{}

\begin{document}

\maketitle

\begin{abstract}
In this short note we propose using agentic swarms of virtual labs as a model of an AI Science Community. In this paradigm, each particle in the swarm represents a complete virtual laboratory instance, enabling collective scientific exploration that mirrors real-world research communities. The framework leverages the inherent properties of swarm intelligence -- decentralized coordination, balanced exploration-exploitation trade-offs, and emergent collective behavior -- to simulate the behavior of a scientific community and potentially accelerate scientific discovery. We discuss architectural considerations, inter-laboratory communication and influence mechanisms including citation-analogous voting systems, fitness function design for quantifying scientific success, anticipated emergent behaviors, mechanisms for preventing lab dominance and preserving diversity, and computational efficiency strategies to enable large swarms exhibiting complex emergent behavior analogous to real-world scientific communities. A working instance of the AI Science Community is currently under development.
\end{abstract}

\begin{figure}[b!]
  \centering
  \includegraphics[width=0.88\textwidth]{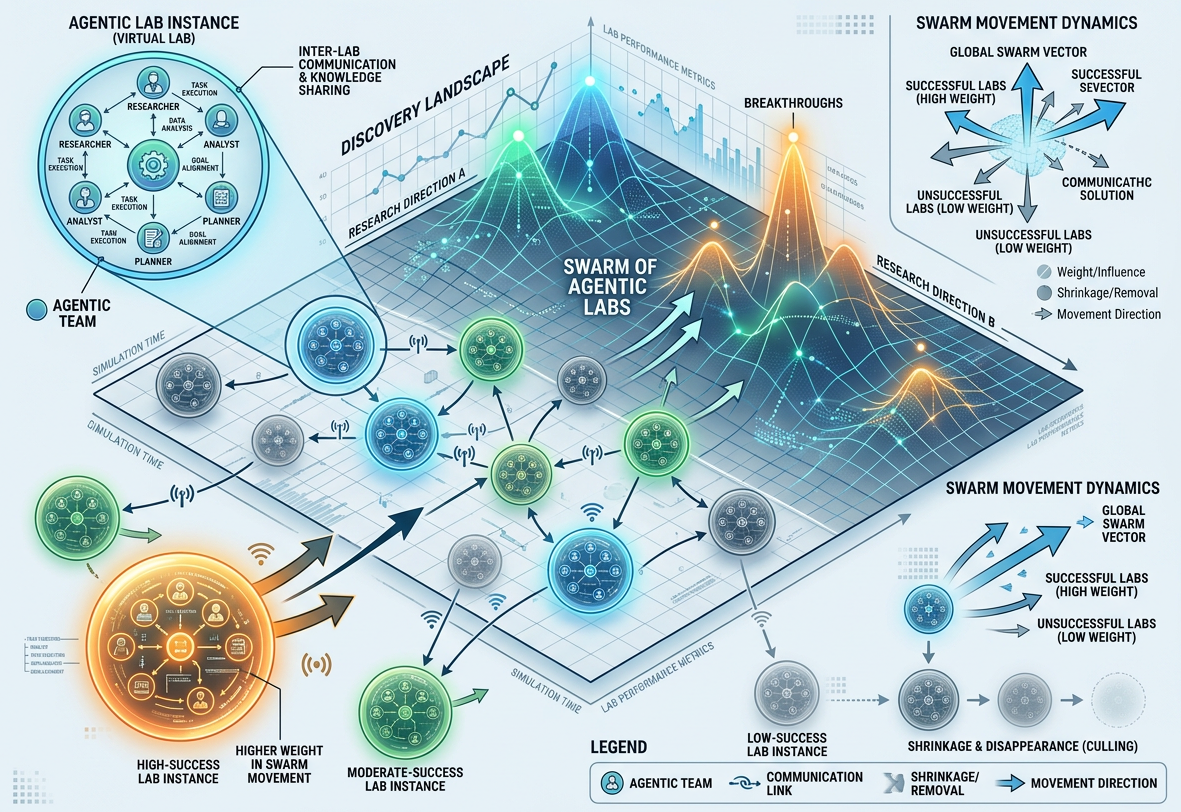}
  \caption{The AI Science Community (generated with Nano Banana 2 at {\tt https://openart.ai/}).}
\end{figure}

\section{Introduction}

The rapid advancement of large language models (LLM) has given rise to agentic {\em virtual labs}, which are autonomous groups of LLM agents \cite{wang2024survey, xi2025rise} capable of performing the full cycle of research, including scanning the literature, formulating hypotheses, trying out solutions, writing a paper, and evaluating its own results \cite{boiko2023autonomous,wu2023autogen,hong2024metagpt,toscano2025athena,lu2024ai, yamada2025ai}.  Here, we propose to extend this paradigm to swarms of virtual labs to mimic the behavior of an entire science community. This {\em AI Science Community} framework combines agentic artificial intelligence with swarm intelligence to create decentralized networks of virtual laboratories. In this novel approach, each particle within the swarm constitutes a complete virtual laboratory instance, enabling collective scientific exploration that mirrors the organizational and behavioral dynamics of real-world research communities. The framework exploits the inherent properties of swarm intelligence---decentralized coordination, balanced exploration-exploitation trade-offs, and emergent collective behavior---to simulate and potentially accelerate the process of scientific discovery.  Below, we detail the proposed swarm mechanism, efficiency considerations, challenges in defining appropriate fitness functions, expected emergent behaviors, and conclusions drawn from this framework.


\section{Swarm Mechanism}

We propose creating a swarm where each individual is an independent virtual laboratory instance that can communicate with other instances, modeling the dynamics of real scientific communities \cite{woolley2010evidence}. The proposed swarm mechanism operates on principles that closely parallel those observed in academic research communities: a positive move by a lab influences all the other labs, similar to how influential publications redirect research efforts across actual labs, and unsuccessful labs shrink while successful labs thrive, mimicking natural selection in academic research \cite{kuhn1962structure}.  However, while labs move towards ``hot'' ideas, still some labs explore alternative paradigms to maintain diversity and not stifle innovation.  The actual mechanism of influence we propose is a ``citation'' system based on voting. A virtual lab can cite the work of other labs by means of agents that up-vote those labs, with more votes analogous to more citations indicating greater scientific impact.

Planning agents in each lab control the exploration-exploitation balance through an exploration rate parameter that determines the probability of generating novel approaches versus moving towards successful approaches from other labs. This balance can be adjusted dynamically, with higher exploration rates in early iterations for broad search and lower rates in later stages for convergence. Evaluation agents function as ``anonymous peer reviewers'' across labs.
The swarm should be initialized with diversity to enable broad exploration of the solution space; initially there is a diversity of ``opinions'' and they disagree, corresponding to high variance and low confidence in the prediction. As time goes on, the swarm moves in a decentralized manner towards good solutions.

\section{Efficiency Considerations}

Computational cost is the bottleneck of swarm intelligence \cite{poli2007particle}. Several strategies can be employed to keep costs down while maintaining the effectiveness of the swarm optimization.
One approach is to start with many ``cheaper'' agents with low priors, which are gradually being pruned into fewer, more complex agents constrained by more informed priors. Another strategy involves using a hybrid architecture where certain agents are shared while others remain independent. The virtual labs might consist of just planning agents, and share the other agents, while still keeping lab independence. Additionally, one can restrict the number of labs in existence at any given time, or restrict the number of iterations in the swarm optimization.

\section{The Fitness Function}

Defining the fitness function, which is a requirement of swarm intelligence, is the problem of defining scientific success. The fitness function could be the error with respect to the reference solutions. However, in practical problems, reference solutions may not be available, and here the fitness function is more difficult to formulate. A promising alternative involves having agents that up-vote the labs, so the labs with more votes -- here is an analogy with more citations -- are the most successful. The peer-review agents could serve as the voting mechanism, assessing outputs from multiple labs and assigning votes based on solution quality, mirroring how real scientific communities operate \cite{merton1973sociology}. In addition, swarms can do multi-objective optimization, so that, in addition to the error, one could also consider computational complexity, convergence, and other metrics \cite{coello2007evolutionary}. 
Practical challenges when reference solutions are unavailable include preventing groupthink where labs reinforce each other's errors \cite{janis1972victims}, and establishing meaningful proxy metrics that correlate with true solution quality without knowing the solution itself.

\section{Expected Emergent Behaviors}

Several types of emergent behaviors are anticipated in virtual lab communities operating under this framework.
Decentralized convergence and consensus formation should emerge, where labs start with a diversity of ``opinions'' and disagreement corresponding to high variance and low confidence, then move in a decentralized manner towards correct solutions. Natural selection of research directions should occur as unsuccessful labs shrink while successful labs thrive, mimicking the dynamics of real scientific communities.
With multiple objectives, the community can split into ``rival camps'' with diverse approaches, with labs clustering around different perspectives similar to how real scientific communities form competing paradigms \cite{kuhn1962structure}. Balanced exploration-exploitation should emerge as labs move towards ``hot'' ideas while still some labs explore other alternative paradigms.

\section{Implementation Outline}
\label{sec:implementation}

Each virtual lab can be implemented as a containerized process using one or more LLM APIs. The various agents are defined by {\em role prompts} and tool permissions. Each lab must have planning capability to decide what to do next and how strongly to explore versus exploit, worker capability to generate hypotheses, designs, code, and written artifacts, and evaluation capability to review the work of other labs in an anonymous peer-review fashion. These roles can be implemented by using the same underlying model or by mixing models. This increases diversity of personality among the agents and allows one to reduce costs by using smaller models for routine steps and stronger models for high-stakes decisions. For safety and robustness,
tool execution should be sandboxed and network access should be constrained as appropriate.

To support decentralized coordination, a global {\em swarm registry} contains a public record of what each lab claims, what evidence it has produced, and how much the community currently trusts it. This information is {\em embedded} optimally into a numerical representation in a high-dimensional vector space. At each iteration, a numerical velocity vector is determined by three factors: its own velocity inertia at the current iteration, its best-known position in the search-space, and the entire swarm’s best-known position. This update depends on social and cognitive coefficients that are decoded from the swarm collective state. The lab position is then updated by adding the velocity vector to its current state.

\section{Related Work}

Swarm intelligence has established itself as a powerful paradigm for decentralized optimization, operating on the principle that there is no central command yet the collective is coordinated \cite{kennedy1995particle, bonabeau1999swarm}. This behavior is observed in animal groups, like flocks of birds or fish schools, but also in human groups, as in research communities. Prior work has demonstrated the effectiveness of swarm-based approaches in computational domains. For example, the PSO-PINN framework integrates Particle Swarm Optimization with Physics-Informed Neural Networks, where each individual in the swarm is a PINN \cite{davi2022pso}. Swarms of LLMs exist, and swarm intelligence has been used to optimize LLM prompts, using evolutionary and particle swarm methods to discover effective prompt formulations \cite{guo2024connecting, fernando2023promptbreeder}. Swarms of LLM agents have been explored for collaborative problem-solving \cite{zhuge2024gptswarm}, but these are individual LLM agents, not swarms of virtual labs. Evolutionary optimization has demonstrated remarkable success in automatically discovering effective model merging recipes for foundation models \cite{akiba2025evolutionary}. On the other hand, ``model soups'' --- weighted averages of multiple fine-tuned models --- can improve accuracy without increasing inference costs, providing a foundation for more sophisticated merging strategies \cite{wortsman2022model}. These evolutionary approaches to model merging align conceptually with swarm intelligence principles, as both paradigms leverage population-based search, decentralized optimization, and emergent collective behaviors to navigate complex solution spaces and discover solutions that may be difficult for human experts to identify manually.

\section{Conclusions}

This note proposes an AI Science Community framework combining agentic AI with swarm intelligence to create decentralized networks of virtual laboratories that simulate scientific communities. Using virtual labs as the particles is interesting because the swarm of virtual labs functions as a simulation of a scientific community, the same way that a single virtual lab stands in as a simulation of a real lab. Some labs are successful and grow; others disappear. The more credible labs have more weight in the movement of the entire community. With multiple objectives, the community can split into ``rival camps'' with diverse approaches. 
A working instance of the AI Science Community is currently under development.

\bibliographystyle{ieeetr}
\bibliography{refs}

\end{document}